\documentclass{article}
\usepackage{spconf,amsmath,graphicx}

\usepackage{dblfloatfix, hyperref}
\usepackage{amsfonts}
\usepackage{booktabs}
\usepackage{enumitem}
\setlist{nosep, leftmargin=14pt}

\usepackage{xcolor}

\hypersetup{
    colorlinks=true,
    linkcolor=blue,
    filecolor=magenta,      
    urlcolor=cyan,
    citecolor=blue,
}

\title{Curriculum-Guided Myocardial Scar Segmentation for \\ Ischemic and Non-ischemic Cardiomyopathy}

\name{\begin{tabular}{c} 
      Nivetha Jayakumar$^{1*}$ \qquad Jonathan Pan$^{2}$ \qquad Shuo Wang$^{2}$ \qquad Bishow Paudel$^{2}$ \\ 
      \textit{Nisha Hosadurg}$^{2}$ \qquad \textit{Cristiane C. Singulane} $^{2}$ \qquad \textit{Sivam Bhatt}$^{2}$ \\ 
      \textit{Amit R. Patel}$^{2}$ \qquad \textit{Miaomiao Zhang}$^{1,3}$ 
      \end{tabular}}

\address{$^1$ Department of Electrical and Computer Engineering, University of Virginia, USA \\ 
         $^2$ Division of Cardiovascular Medicine, University of Virginia Health System, USA \\ 
         $^3$ Department of Computer Science, University of Virginia, USA}

\begin{document}
\maketitle

Identification and quantification of myocardial scar is important for diagnosis and prognosis of cardiovascular diseases. However, reliable scar segmentation from Late Gadolinium Enhancement Cardiac Magnetic Resonance (LGE-CMR) images remains a challenge due to variations in contrast enhancement across patients, suboptimal imaging conditions such as post-contrast washout, and inconsistencies in ground-truth annotations on diffuse scars caused by inter-observer variability. In this work, we propose a curriculum learning–based framework designed to improve segmentation performance under these challenging conditions. The method introduces a progressive training strategy that guides the model from high-confidence, clearly defined scar regions to low-confidence or visually ambiguous samples with limited scar burden. By structuring the learning process in this manner, the network develops robustness to uncertain labels and subtle scar appearances that are often underrepresented in conventional training pipelines. Experimental results show that the proposed approach enhances segmentation accuracy and consistency, particularly for cases with minimal or diffuse scar, outperforming standard training baselines. This strategy provides a principled way to leverage imperfect data for improved myocardial scar quantification in clinical applications. Our code is publicly available on \href{https://github.com/vfb8zv/Curriculum-Guided-Myocardial-Scar-Segmentation-for-Ischemic-and-Non-ischemic-Cardiomyopathy/}{Github}.

\section{Introduction}
Accurate segmentation and quantification of myocardial scar from LGE-CMR images is critical for assessing disease severity and guiding therapy in cardiomyopathies~\cite{xing2025scar}. The extent and spatial distribution of scar correlate with arrhythmic risk and treatment response, yet consistent delineation remains difficult due to substantial variations in image contrast, diffuse enhancement patterns, and inter-observer related annotation uncertainty~\cite{kellman2012cardiac}. In ischemic cardiomyopathy, detecting infarcted myocardium can help guide revascularization procedures whereas in non-ischemic cases, quantifying scar burden can serve as a key diagnostic and prognostic tool~\cite{hunold2005myocardial}.

Recent advances in deep learning have enabled strong performance in LGE-CMR segmentation, achieving near-human accuracy under ideal conditions with good image quality and relatively large, well-defined scars~\cite{tavakoli2025scarnet,jayakumar2025deep}. However, these models struggle in clinically important but visually challenging cases, such as low-contrast enhancement, diffuse scar, and regions affected by post contrast washout \cite{karim2016evaluation}. As a result, the distinction between scar and viable myocardium becomes ambiguous and ground-truth labels show high interobserver variability. Existing supervised approaches assume clean annotations and matched contrast distributions, making them fail under realistic imaging conditions~\cite{karim2016evaluation}. Moreover, current datasets are dominated by well-contrasted examples, biasing models toward easy cases and leaving failure cases unexplored. As a result, substantial drops in segmentation performance are observed for difficult samples~\cite{moafi2025robust}, thus emphasizing the need for learning paradigms that explicitly address uncertainty in visual appearance and annotation variance. 

Curriculum learning has been shown to improve model robustness in scenarios with noisy or uncertain labels by introducing samples in order of increased difficulty~\cite{zhou2020robust}. This approach enables the network to learn stable representations before seeing ambiguous or low-contrast cases, thus reducing the risk of overfitting to noisy data. In this work, we propose a curriculum learning–based framework for myocardial scar segmentation that progressively trains the model from high-confidence, well-defined scars to ambiguous, low-contrast regions. Our contributions are threefold:
\begin{enumerate}
    \item A robust scar segmentation framework explicitly designed to handle low-contrast, diffuse myocardial fibrosis.
    \item A dynamic curriculum-guided training paradigm with a progressive, difficulty-aware schedule to increase robustness to uncertain labels.
    \item A hybrid loss function with increased focus on smaller scar regions through a foreground-weighted Dice metric. 
\end{enumerate}
 Experimental results on 2D multi-center LGE-CMR datasets demonstrate our framework's improved segmentation accuracy and stability, notably for diffuse or low-contrast scars.

\begin{figure*}[!htp]
\centering
\centerline{\includegraphics[width=0.95\linewidth]{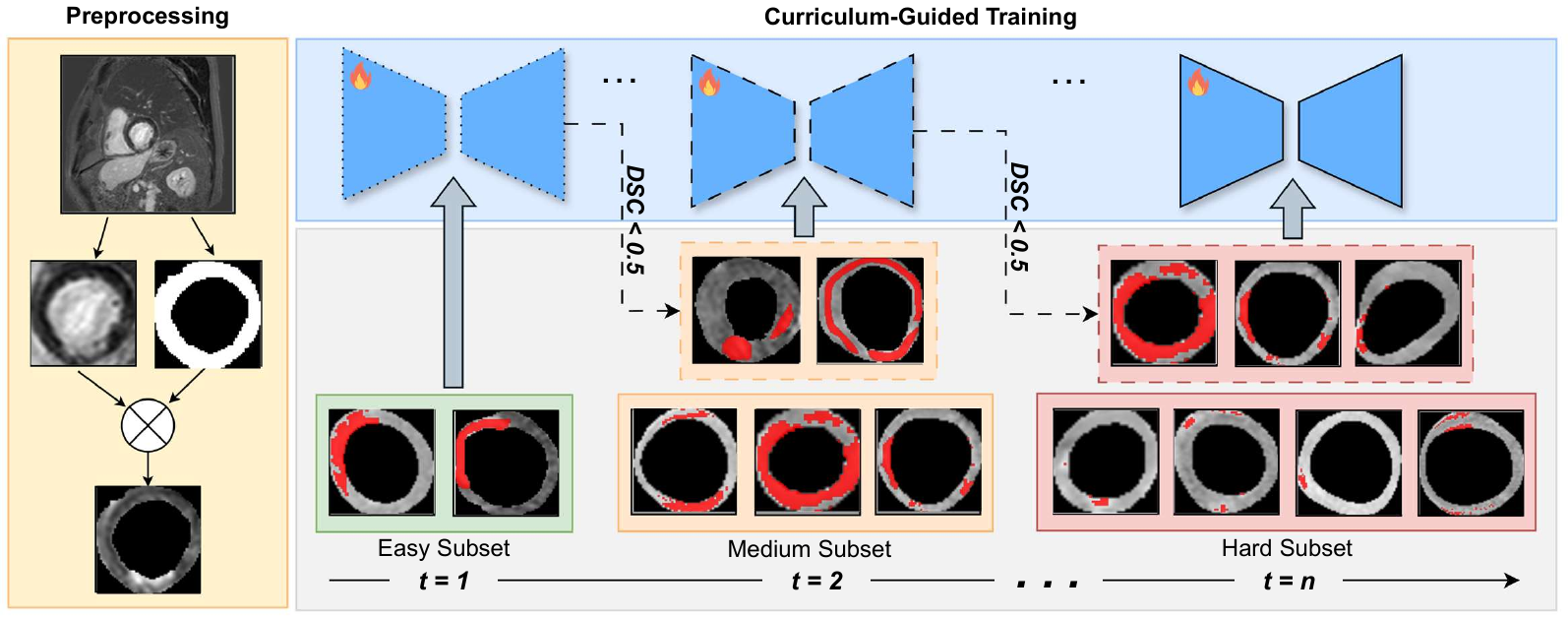}}
\caption{Curriculum-guided training framework for LGE segmentation, using an arbitrary segmentation backbone, integrating implicit and explicit difficulty-aware labels.}
\label{fig:main_arch}
\end{figure*} 

\section{Methodology}
Consider a training dataset of LGE-CMR images $\{ I_i\}_{i=1}^N$ with corresponding ground-truth scar labels $\{ Y_i\}_{i=1}^N$ where $Y_i$=1(scar) and 1-$Y_i$(non-scar), left ventricular (LV) myocardium contours $\{M_i\}_{i=1}^N$, with $N$ denoting the total number of images. Note that the LV contours can be manually labeled or predicted by DL models~\cite{xing2023joint}. Our goal is to learn a segmentation model that can robustly differentiate myocardial scar from masked myocardium by $M_i$ under heterogeneous imaging conditions which include low-contrast images, diffuse scar and labels with high inter-observer variance. \\
\noindent{\bf Curriculum-Guided Segmentation Network.} Our framework incorporates a segmentation backbone trained using a difficulty-aware sampling strategy. Each image $I_i$ is assigned a difficulty score, $d_i \in [0,2]$ (easy, medium, and hard) determined by clinical experts based on the LGE-CMR intensity contrast and inter-observer variability. Based on these scores, the dataset is partitioned into subsets $\mathcal{D}_{d_i \in [0,2]}$, forming the basis for the staged training.

The training process consists of a number of $n$ stages corresponding to progressively increasing levels of data difficulty. It begins with a subset of “easy” samples, followed by gradually incorporating more difficult cases as the model's training stabilizes. 
At each stage $t \in [1,\cdots,n]$, the network parameters $\theta_t$ are updated using the 
cumulative dataset, thus ensuring that the network first learns robust scar representations from clear, high-contrast examples before adapting to ambiguous regions. In addition to the supervised curriculum established by clinicians, we introduce an adaptive mechanism that allows the model to reselect difficult samples during training. Specifically, at the end of each stage $t$, samples leading to spikes in the loss curve are re-introduced in the next stage $t+1$. This allows the model to dynamically refine its notion of sample difficulty based on its learning trajectory.

Formally, the segmentation model, $f_\theta(\cdot)$, predicts a scar 
probability map, $\hat{Y_i} = f_\theta(I_i)$. The training objective is applied 
iteratively across curriculum stages with the cumulative dataset, i.e., 
{\small
\begin{equation}
\label{eq:objective}
\theta_{t+1} = \arg\min_{\theta} \mathbb{E}_{(I_t, Y_t) \in \mathcal{D}_t} \mathcal{L}(Y_t,f_\theta(I_t)) + \lambda \text{Reg}(\theta),
\end{equation}}
 \noindent where $\mathcal{L}$ represents a loss function with $\text{Reg}(\cdot)$ denoting a network regularizer, and $\lambda$ is a weighting parameter. This gradual learning process stabilizes optimization and prevents the model 
from overfitting to noisy or ambiguous labels in the early stages of training. To evaluate the generalizability of the proposed framework, we implemented it across three different segmentation architectures~\cite{oktay2018attention,chen2021transunet,tavakoli2025scarnet}.

The loss function in Eq.~\eqref{eq:objective} is defined as a weighted combination of a foreground-weighted Dice loss~\cite{milletari2016v} and a focal loss~\cite{lin2017focal}, i.e., $\mathcal{L} =  \mathcal{L}_{\text{wDice}} + \beta \mathcal{L}_{F}$, where $\beta$ is a weighting parameter between the two loss components. In particular, the weighted Dice loss focuses more on correct segmentation within the myocardial region while simultaneously accounting for background pixels. We can formulate it as
{
\small
\begin{equation}
\mathcal{L}_{\text{wDice}} = 1 - \Big( w_{\text{fg}} \cdot \text{Dice}_{\text{fg}} + w_{\text{bg}} \cdot \text{Dice}_{\text{bg}} \Big),
\end{equation}}
\noindent where $\text{Dice}_{\text{fg}} = \frac{2 \sum_{i \in {M}} Y_i \hat{Y}_i + \epsilon}{\sum_{i \in {M}} Y_i + \sum_{i \in {M}} \hat{Y}_i + \epsilon}$, and $\text{Dice}_{\text{bg}}$ is defined similarly with setting $Y_i \rightarrow (1 - Y_i), \hat{Y}_i \rightarrow (1 - \hat{Y}_i)$. Here, $w_{\text{fg}}, w_{\text{bg}}$ are weights to balance the extent of overlap in foreground and background pixels defined as $\text{Dice}_{\text{fg}}$ and $\text{Dice}_{\text{bg}}$, and $\epsilon$ is a constant defined for numerical stability. 

The focal loss~\cite{lin2017focal} is employed to to further penalize 
\begin{figure*}[!hb]
\centering
\centerline{\includegraphics[width=1\linewidth]{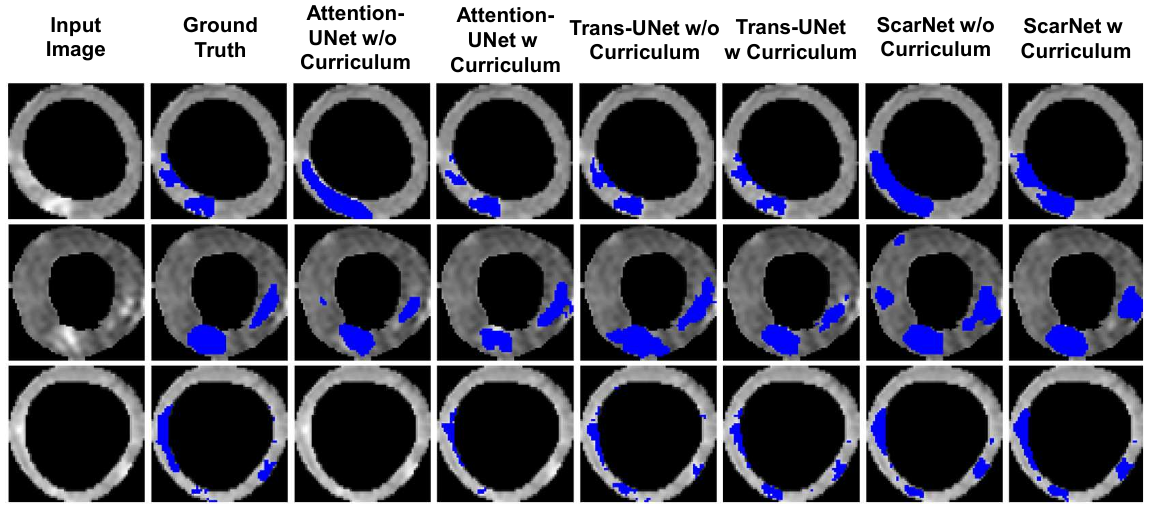}}
\caption{ Comparison of predicted scar segmentation with and without curriculum guidance. Left to right: input images overlayed with ground truth LGE annotations, AttentionUNet, TransUNet and ScarNet trained with and without curriculum guidance.}
\label{fig:res_viz}
\end{figure*} 
false negatives in small scar regions, i.e., 
{\small
\begin{equation}
\mathcal{L}_F =
- \alpha  Y \cdot (1 - \hat{Y})^\gamma  \log(\hat{Y})
- (1 - \alpha)\\(1 - Y) \cdot \hat{Y}^\gamma  \log(1 - \hat{Y})
\end{equation}}

\noindent where $\alpha$ balances false positives and negatives, $\gamma$ is a scalar weighting parameter emphasizing hard-to-classify pixels.

\section{Experimental Results}

\begin{figure*}[!htbp]
\centering
\centerline{\includegraphics[width=1.0\linewidth]{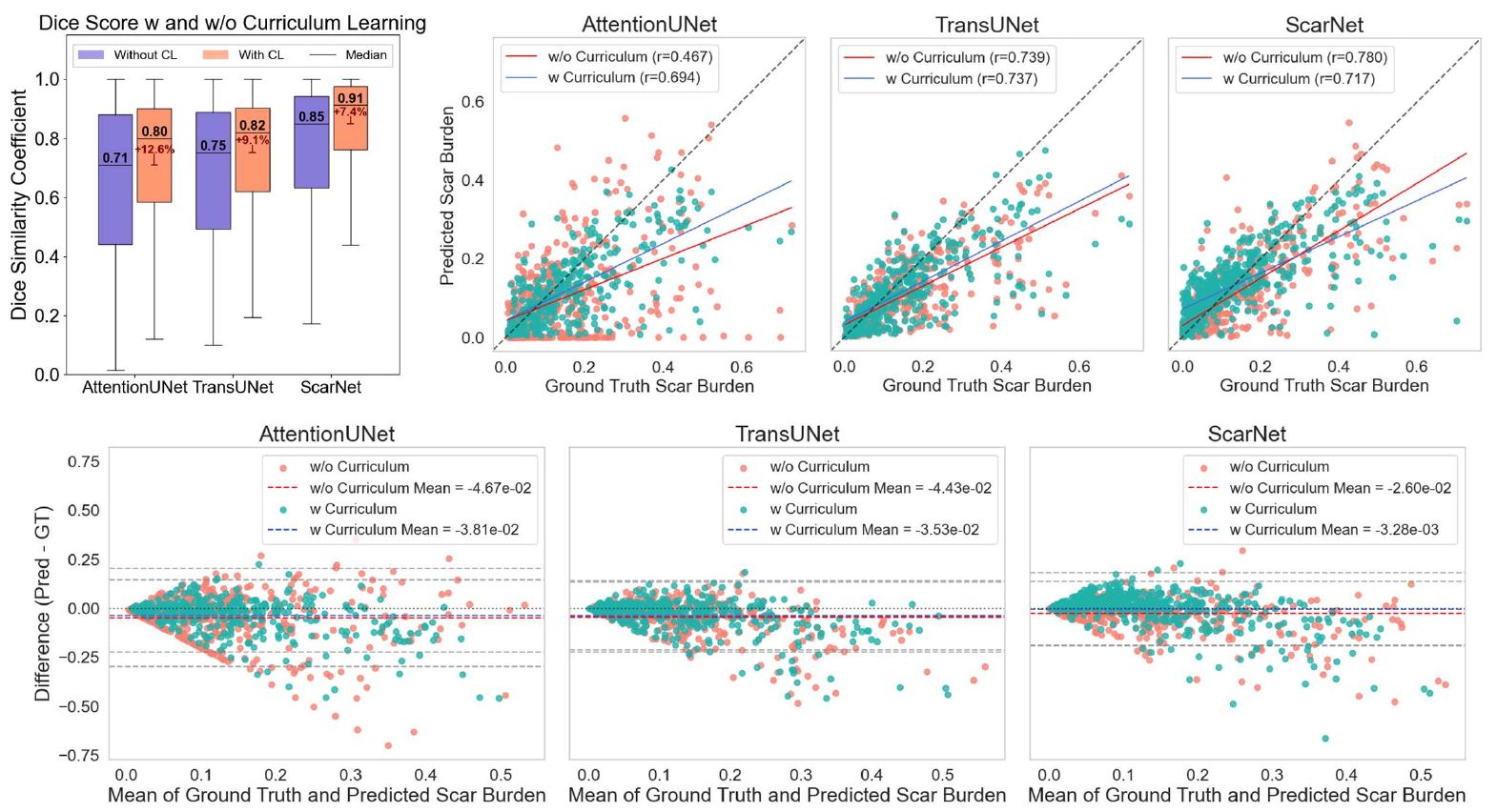}}
\caption{Comparison of (a) Dice similarity coefficients, (b) Pearson Correlation and (c) Bland Altman plots for LGE segmentation and scar burden across 3 backbones (AttentionUNet, TransUNet, and ScarNet) trained with and without curriculum guidance}
\label{fig:dice_quant}
\end{figure*} 

To test the efficacy of our framework, we use a curated LGE-CMR dataset collected in-house, across 3 different scanners and 2 sites, thus comprising of patients with varying degrees of myocardial scarring and image quality. To assess the generalizability of the framework, we compare three different segmentation backbones and compare it against a standard supervised training strategy. To evaluate the extent of overlap between the ground-truth and predicted segmentation, we compute the Dice similarity coefficient \cite{dice1945measures} as $
\text{Dice}(Y, \hat{Y}) = 
\frac{
2 \sum_{i \in {Y}} Y_i \hat{Y}_i
}{
\sum_{i \in {Y}} Y_i + \sum_{i \in {Y}} \hat{Y}_i
}
$. We also compute the scar burden from the ground truth and predicted scar regions as $\frac{\sum_{i \in {M}} y_i}{\sum_{i \in {M}} (M_i>0)}, \text{ where } y\in \{\hat{Y}, Y\} \text{ and } M \text{ is the myocardium}$. We use this metric in Bland–Altman plots to assess the degree of over- or underestimation of the scar, as well as its linear correlation with the increase in scar area. Clinically, scar burden is an important biomarker for identifying and differentiating specific types of cardiomyopathy.

\subsection{Experimental Settings}
\noindent\textbf{Dataset.} We used LGE-CMR images from $875$ patients \cite{jayakumar2025deep}, a mixed cohort with both ischemic and non-ischemic cardiomyopathy, with manually annotated regions of myocardial scars, endocardial and epicardial boundaries semi-automatically labelled using suiteHEART~\cite{suiteheart}. A total of $4065$ 2D slices with regions of LGE with corresponding scar annotations were extracted for all patients, followed by cropping and masking them to the left-ventricular myocardium (ROI) using the provided contours. The slices were resized to $64^2$. We used $80$\% of this processed data for training and $20$\% as an independent test set. We had a total of $t=3$ levels of difficulty for each 2D slice used for progressive curriculum learning. All images and labels were collected, processed and provided by 5 CMR experts, where each subject had a single expert annotator.

\noindent\textbf{Implementation Details.} The networks are trained using the Adam optimizer \cite{kingma2014adam} with a learning rate of $1e-4$ and weight decay $\in [1e\text{-}9, 1e\text{-}14]$. We train with a batch size of $32$ for $t=1$ and reduce it to $4$ for $t=2,3$, thus allowing the model to gradually learn the ambiguous samples. The optimal empirical hyper-parameters are $\alpha = 0.25, \beta =0.5, \gamma=2, w_{\text{fg}}=(0.6, 0.75, 0.8) \text{ and } w_{\text{bg}}= (0.4, 0.25, 0.2)$ for $t=1,2,3$ respectively. The total training epochs range from $800$ to $1500$ for the different backbones. All experiments are conducted on NVIDIA A40 and RTX-A6000 GPUs. 

\noindent\textbf{Results.} Fig.~\ref{fig:res_viz} presents examples of LGE-CMRs with manually labeled myocardial scar annotations and predicted scar for three different backbones, AttentionUNet~\cite{oktay2018attention}, TransUNet \cite{chen2021transunet} and ScarNet \cite{tavakoli2025scarnet}. The examples show that our curriculum-guided training framework is robust in detecting disjoint, sparse and diffuse myocardial scar as compared to the conventional supervised training approach. Examples in the third row also demonstrate improved delineation in low-contrast regions, where AttentionUNet fails to detect any scar with the standard training approach, and TransUNet underestimates the total scar region.

\begin{table}[!h]
\centering
\resizebox{\columnwidth}{!}{%
\begin{tabular}{lccc}
\toprule
\textbf{p-values} & \textbf{AttentionUNet} & \textbf{TransUNet} & \textbf{ScarNet} \\
\midrule
Dice Score & $7.60\times10^{-5}$ & $2.24\times10^{-6}$ & $7.25\times10^{-31}$ \\
Scar Burden & $1.67\times10^{-7}$ & $1.30\times10^{-2}$ & $1.85\times10^{-4}$ \\
\bottomrule
\end{tabular}
}
\caption{P-values for Dice Scores and Absolute Scar Burden Error for 3 models with and without curriculum guidance.}
\label{tab:pvalues}
\end{table}

Fig.~\ref{fig:dice_quant} provides plots for quantitative evaluation metrics, including the Dice to indicate the extent of overlap between the ground truth and predicted scar regions, the Pearson correlation plots~\cite{bravais1844analyse} between the ground truth and predicted scar burdens showing the linear agreement between the two variables and the model's performance with increasing scar area, followed by the Bland Altman plots~\cite{altman1983measurement} indicating the difference between the ground truth and predicted scar burdens along with the extent of over and under-estimated segmentation maps. Additionally, we performed the paired Wilcoxon signed-rank test~\cite{rosner2006wilcoxon} to evaluate statistically significant improvements in segmentation performance with and without curriculum guidance. The computed p-values reported in table~\ref{tab:pvalues} indicate that all three architectures show significant differences in both Dice score and absolute scar burden error. Moreover, the best performing model, ScarNet, achieved a $32.6\%$ increase in median Dice from $0.49$ to $0.65$ on 6 known difficult cases. Our method demonstrates robust performance, achieving a higher median DSC across all backbones and lower mean difference between the ground truth and predicted scar areas, thus indicating a higher correlation and lower scar burden.

\section{Conclusion}
This paper presents a curriculum-guided segmentation framework for robust myocardial scar detection from LGE-CMR with noisy, low confidence labels. Our approach employs a progressive training strategy that guides the model gradually learning from high to low confidence samples. This curriculum-based approach enables the model to learn the scar structure and gradually generalize to label noise, intensity variability, and diffuse myocardial enhancement patterns that typically challenge the standard supervised training methods. Experimental results demonstrate that the Curriculum-Guided segmentation framework in conjunction with the weighted loss that emphasizes scar patterns and penalizes hard-to-classify pixels, achieves higher segmentation accuracy and consistency, particularly in difficult cases with minimal or heterogeneous scar appearance. The experimental results show that this work provides improved reliability of myocardial scar quantification in clinical settings by achieving higher accuracy in detecting regions of scar.

\noindent\textbf{Acknowledgments} - This work was supported by NIH 1R21EB032597, iPRIME Student Fellowship Award, GEHealthcare, Area19, CircleCVI, Neosoft and Siemens Healthineers. \\
\textbf{Compliance with Ethical Standards} - All studies involving human subjects and waiver of consent were approved by our institutional review board.


\bibliographystyle{IEEEbib}
\bibliography{strings,refs}

\end{document}